\documentclass[pmlr]{jmlr}

\usepackage{longtable}

\usepackage{booktabs}
\usepackage[load-configurations=version-1]{siunitx} 

\usepackage{graphicx}
\graphicspath{ {./images/} }

\theorembodyfont{\upshape}
\theoremheaderfont{\scshape}
\theorempostheader{:}
\theoremsep{\newline}

\jmlrvolume{1}
\jmlryear{2021}
\jmlrworkshop{NeurIPS 2021 Competition and Demonstration Track}

\title{MEWS: Real-time Social Media Manipulation Detection and Analysis}

 \author{
 \Name{Trenton W. Ford} \Email{tford5@nd.edu}\\
 \Name{Michael Yankoski} \Email{myankosk@nd.edu}\\
 \Name{William Theisen} \Email{wtheisen@nd.edu}\\
 \Name{Tom Henry} \Email{thenry3@nd.edu}\\
 \Name{Farah Khashman} \Email{fkhashma@nd.edu}\\
 \Name{Katherine R. Dearstyne} \Email{kdearsty@nd.edu}\\
 \Name{Tim Weninger} \Email{tweninger@nd.edu}\\
 \addr 
  Department of Computer Science and Engineering\\
  University of Notre Dame\\
  Notre Dame, IN 46530 \\
 \AND
 \Name{Pamela Bilo Thomas} \Email{pamela.thomas.1@louisville.edu}\\
 \addr 
 Department of Computer Science and Engineering\\
 University of Louisville\\
 Louisville, KY 40292\\
}

  \editors{Douwe Kiela, Marco Ciccone, Barbara Caputo}

\begin{document}

\maketitle

\begin{abstract}
This article presents a beta-version of MEWS (Misinformation Early Warning System). It describes the various aspects of the ingestion, manipulation detection, and graphing algorithms employed to determine--in near real-time--the relationships between social media images as they emerge and spread on social media platforms. By combining these various technologies into a single processing pipeline, MEWS can identify manipulated media items as they arise and identify when these particular items begin trending on individual social media platforms or even across multiple platforms. The emergence of a novel manipulation followed by rapid diffusion of the manipulated content suggests a disinformation campaign.
\end{abstract}
\begin{keywords}
Social media, misinformation, graph theory, near real-time detection
\end{keywords}

\section{Introduction}
\label{sec:intro}

\begin{figure}[h]
    \centering
    \includegraphics[width=7.2cm]{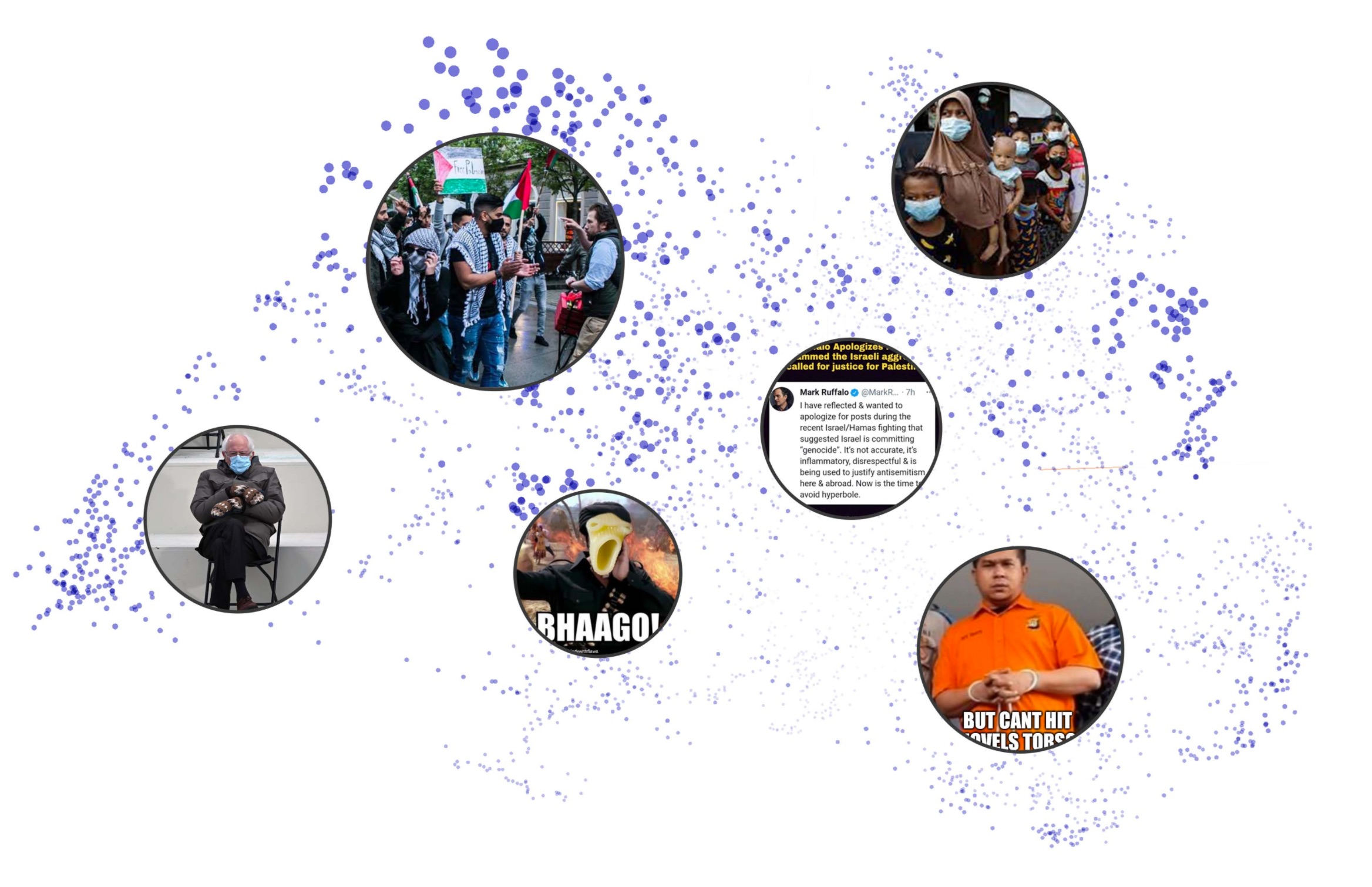} \quad
    \includegraphics[width=7.2cm]{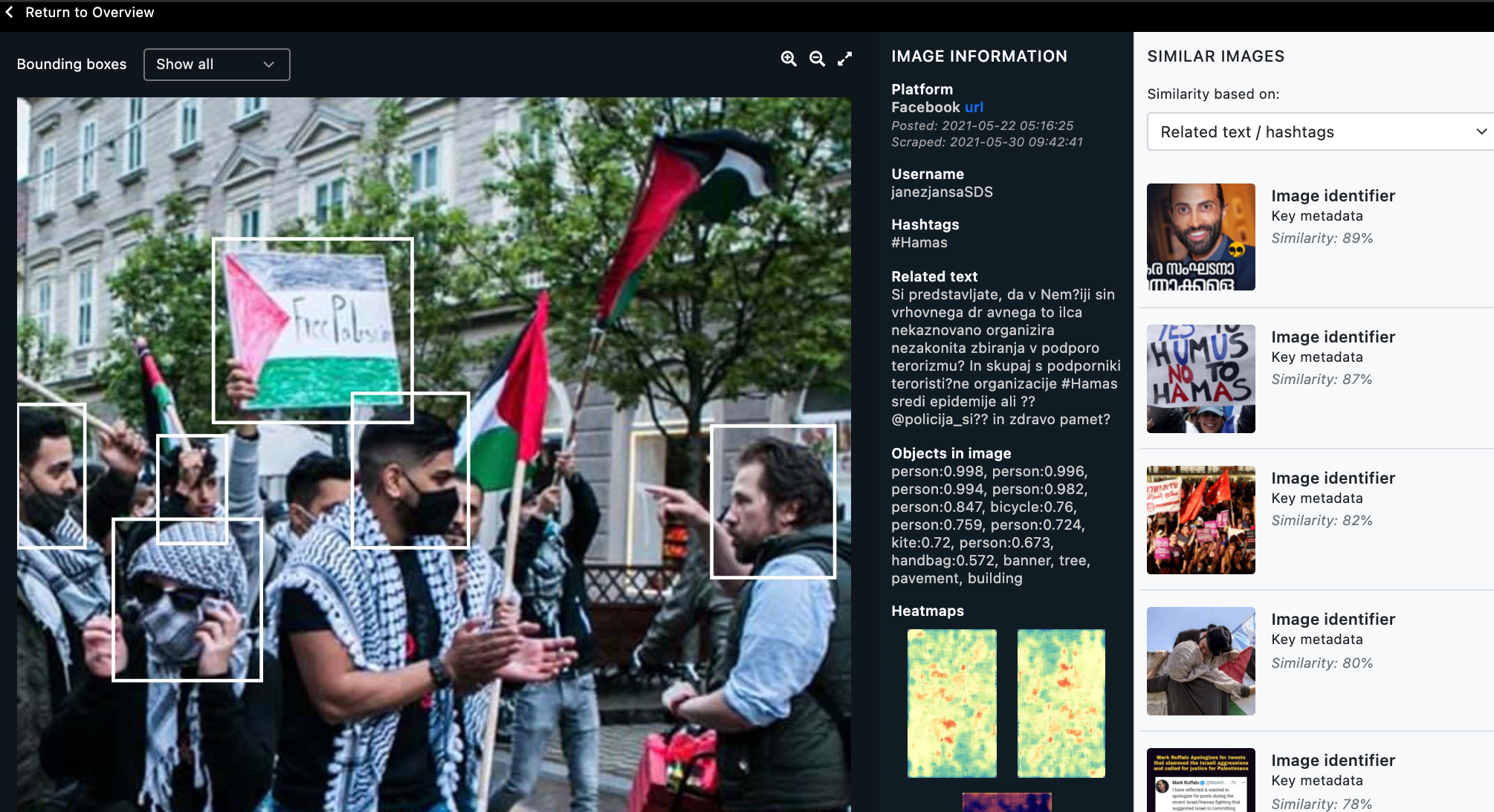}
    \caption{(Left) The macro-view of MEWS displaying image clusters and image-representatives of each cluster. (Right) The image detail page of MEWS including embedded results of image forensics, facial recognition, meme-text, related images and other pertinent information.}
    \label{fig:img1}
\end{figure}  
    
One of the most challenging aspects of online disinformation is the overwhelming volume of content published on social media platforms. Typical approaches to combating this problem rely on tracking headlines, news stories, hashtags, and accounts~\cite{ferrara2017disinformation,glenski2018propagation}. However, disinformation campaigns are becoming increasingly visual~\cite{theisen2020automatic}, but organizing and analyzing this volume of content in the hope of detecting disinformation campaigns in near real-time is impossible for humans without the assistance of automated tools. This problem is especially pertinent in young and struggling democracies whose traditional media organizations lack the ability to keep pace with the explosion of deep-fake, manipulated, altered, or plainly-fake online media~\cite{yankoski2021meme}. To provide such capacity, we have developed a real-time social media manipulation detection and analysis system called MEWS (Misinformation Early Warning System)~\cite{yankoski2020ai}. Figure~\ref{fig:img1} shows MEWS' similarity graph with top images present (on the left) and a single image's detail page (on the right). 

This system combines work in digital forensics, computer vision, graph analysis, and media studies to accomplish three specific tasks: 
\begin{enumerate}
    \item MEWS ingests enormous amounts of images and video from various social media platforms (\textit{e.g.}, Facebook, Instagram, Twitter, Telegram) using keyword targets provided by partner media organizations with context-specific knowledge domains from across the world;
    \item MEWS employs state-of-the-art AI systems to detect and extract faces~\cite{albiero2021img2pose,guo2021sample}, objects~\cite{carion2020end}, text (including meme-text)~\cite{lee2012improving}, image features~\cite{bay2006surf}, and any potential manipulations~\cite{bianchi2011improved} from the visual content; and 
    \item  MEWS constructs a media-graph which pairs similar sub-images, objects, and manipulations for display in an interactive, easily-navigable, and searchable user interface.
\end{enumerate}

For NeurIPS 2021 we demonstrated MEWS' organizational and analytic capabilities using tens of millions of images (and other media) collected from several social media platforms (Facebook, Instagram, and Twitter) from within the Indonesian social-media context. In particular, this demonstration highlighted MEWS' ability to:

\begin{enumerate}
    \item Interactively reveal emergent trends in social media images in (near) real-time.
    \item Identify media manipulations and alterations that recur across media items and platforms. 
    \item Represent the relationships between social media posts on a variety of axes, including: meme-text, ancillary post-text (\textit{e.g.}, hashtags), detected objects, faces, and their identities, etc. 
    \item Provide a searchable interface that users (\textit{i.e.}, newsrooms, civil society, government agencies, and others) can use to understand the way in which disinformation is spreading within online social media.
    \item Provide authenticated end-users the ability to upload media and visualize their relationships to other media contained in the MEWS dataset.
\end{enumerate}

\section{Examples of MEWS's Capabilities}
\begin{figure}[t]
    \centering
    \includegraphics[width=9.0cm]{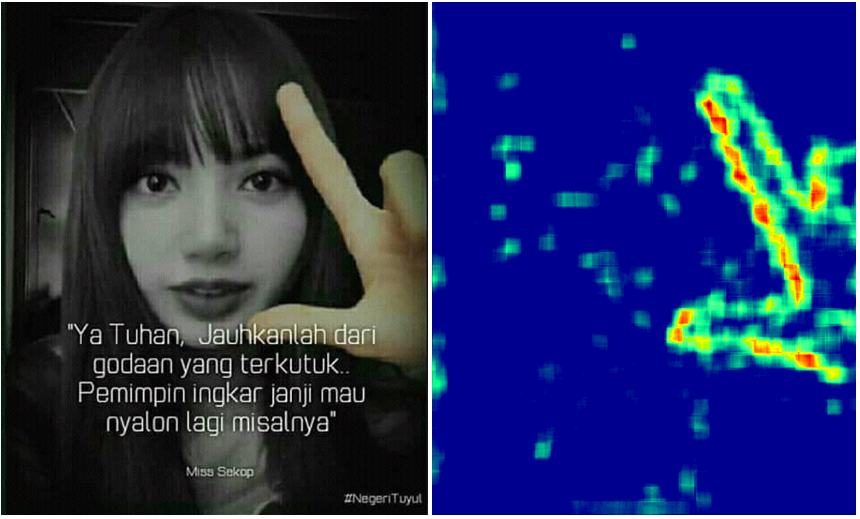} \qquad
    \includegraphics[width=5.2cm]{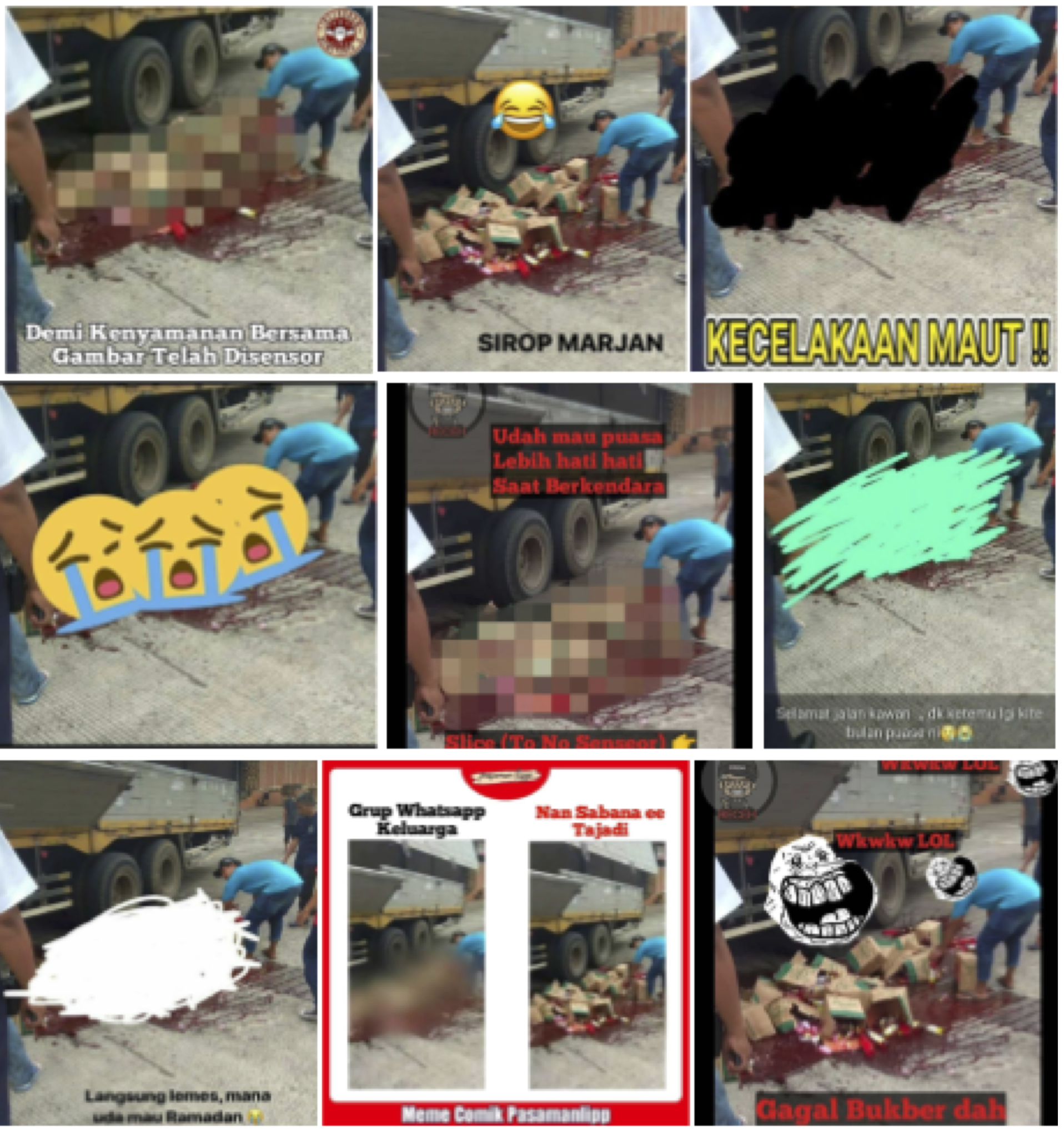}
    \caption{(Left) Example of image manipulation detection and text recognition. (Right) Cluster of self-similar images that represent a possible social media influence campaign.}
    \label{fig:img2}
\end{figure} 
For example, MEWS extracted the manipulated/inserted finger-and-thumb motif from the image in~\figureref{fig:img2} (left) as well as the overlayed text and several other image features. In this particular instance, we find that the finger-and-thumb motif was frequently inserted to show support for a political candidate in Indonesia inauthentically.~\figureref{fig:img2} (right) shows another example of a cluster of altered images of an industrial accident that MEWS detected~\cite{theisen2020automatic}.

\begin{figure}[t]
    \centering
    \includegraphics[width=15cm]{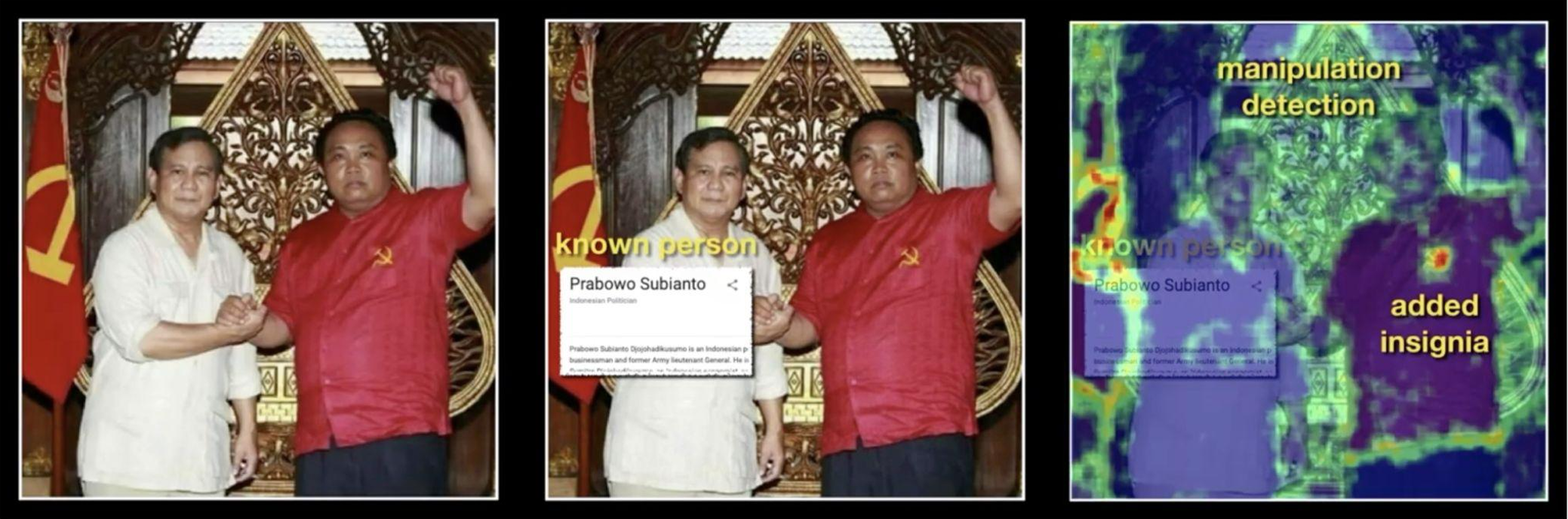} \qquad
    \caption{(Left) The image as it appeared on Indonesian social media. (Center) Facial recognition identified the person on the left as Prabowo Subianto, a presidential candidate. (Right) Manipulation detection algorithms identified areas of high probability image manipulation artifacts.}
    \label{fig:img3}
\end{figure}  
MEWS also identified an image manipulation aimed at political influence:~\figureref{fig:img3}. A 2019 Presidential Candidate is pictured standing with another person whose shirt has been manipulated to display a hammer and sickle. The flag at the left in the image has also been manipulated with similar symbols. The heat map on the right is the algorithmic detection of the image's manipulated portion(s) as identified by the MEWS system. 
    
\section{Interactive Virtual Demonstration}

MEWS provided NeurIPS attendees an opportunity to witness several state-of-the-art technologies applied to a pertinent social problem. During the virtual conference, users were able to browse the existing collection of images through a Google Maps-like interface. They explored the faces, objects, alterations, text extractions, and information MEWS collected from each image. 

In addition to demonstrating the combined power of several AI technologies, MEWS is one of the first image-based social listening and detection services, filling a wide gap in the study of social media, communications, and international security.

\section{The Path Ahead}

Disinformation campaigns will likely continue to influence social, political, and economic processes for the foreseeable future. We believe that these disinformation campaigns will becoming increasingly visual -- taking the style of low-effort memes and out-of-context or cropped photography rather than sophisticated Deep-fakes. 

MEWS provides an early-stage example of the application of AI technologies in the service of helping human users better navigate their social media networks. Despite its performance capabilities, MEWS is not intended to be a standalone solution. Rather, we envision MEWS as a tool for use by a robust partner network of fact-checkers, journalists, human rights watchers, and potentially government representatives who will use the information MEWS surfaces to identify and respond to disinformation threats more efficiently as they emerge on social media.

\acks{This work was supported by the US Agency for International Development (USAID) Cooperative Agreement number 7200AA18CA00059 and by the Defense Advanced Research Projects Agency and the Air Force Research Laboratory under agreement number FA8750-16-2-0173.}

\bibliography{ref}

\begin{thebibliography}{11}
\providecommand{\natexlab}[1]{#1}
\providecommand{\url}[1]{\texttt{#1}}
\expandafter\ifx\csname urlstyle\endcsname\relax
  \providecommand{\doi}[1]{doi: #1}\else
  \providecommand{\doi}{doi: \begingroup \urlstyle{rm}\Url}\fi

\bibitem[Albiero et~al.(2021)Albiero, Chen, Yin, Pang, and
  Hassner]{albiero2021img2pose}
Vítor Albiero, Xingyu Chen, Xi~Yin, Guan Pang, and Tal Hassner.
\newblock img2pose: Face alignment and detection via 6dof, face pose
  estimation.
\newblock In \emph{CVPR}, 2021.
\newblock URL \url{https://arxiv.org/abs/2012.07791}.

\bibitem[Bay et~al.(2006)Bay, Tuytelaars, and Van~Gool]{bay2006surf}
Herbert Bay, Tinne Tuytelaars, and Luc Van~Gool.
\newblock Surf: Speeded up robust features.
\newblock In \emph{European conference on computer vision}, pages 404--417.
  Springer, 2006.

\bibitem[Bianchi et~al.(2011)Bianchi, De~Rosa, and Piva]{bianchi2011improved}
Tiziano Bianchi, Alessia De~Rosa, and Alessandro Piva.
\newblock Improved dct coefficient analysis for forgery localization in jpeg
  images.
\newblock In \emph{2011 IEEE International Conference on Acoustics, Speech and
  Signal Processing (ICASSP)}, pages 2444--2447. IEEE, 2011.

\bibitem[Carion et~al.(2020)Carion, Massa, Synnaeve, Usunier, Kirillov, and
  Zagoruyko]{carion2020end}
Nicolas Carion, Francisco Massa, Gabriel Synnaeve, Nicolas Usunier, Alexander
  Kirillov, and Sergey Zagoruyko.
\newblock End-to-end object detection with transformers.
\newblock In \emph{European Conference on Computer Vision}, pages 213--229.
  Springer, 2020.

\bibitem[Ferrara(2017)]{ferrara2017disinformation}
Emilio Ferrara.
\newblock Disinformation and social bot operations in the run up to the 2017
  french presidential election.
\newblock \emph{arXiv preprint arXiv:1707.00086}, 2017.

\bibitem[Glenski et~al.(2018)Glenski, Weninger, and
  Volkova]{glenski2018propagation}
Maria Glenski, Tim Weninger, and Svitlana Volkova.
\newblock Propagation from deceptive news sources who shares, how much, how
  evenly, and how quickly?
\newblock \emph{IEEE Transactions on Computational Social Systems}, 5\penalty0
  (4):\penalty0 1071--1082, 2018.

\bibitem[Guo et~al.(2021)Guo, Deng, Lattas, and Zafeiriou]{guo2021sample}
Jia Guo, Jiankang Deng, Alexandros Lattas, and Stefanos Zafeiriou.
\newblock Sample and computation redistribution for efficient face detection.
\newblock \emph{arXiv preprint arXiv:2105.04714}, 2021.

\bibitem[Lee and Smith(2012)]{lee2012improving}
Dar-Shyang Lee and Ray Smith.
\newblock Improving book ocr by adaptive language and image models.
\newblock In \emph{2012 10th IAPR International Workshop on Document Analysis
  Systems}, pages 115--119. IEEE, 2012.

\bibitem[Theisen et~al.(2020)Theisen, Brogan, Thomas, Moreira, Phoa, Weninger,
  and Scheirer]{theisen2020automatic}
William Theisen, Joel Brogan, Pamela~Bilo Thomas, Daniel Moreira, Pascal Phoa,
  Tim Weninger, and Walter Scheirer.
\newblock Automatic discovery of political meme genres with diverse
  appearances.
\newblock \emph{Proceedings of the International AAAI Conference on Web and
  Social Media}, 15, 2020.

\bibitem[Yankoski et~al.(2020)Yankoski, Weninger, and Scheirer]{yankoski2020ai}
Michael Yankoski, Tim Weninger, and Walter Scheirer.
\newblock An ai early warning system to monitor online disinformation, stop
  violence, and protect elections.
\newblock \emph{Bulletin of the Atomic Scientists}, 76\penalty0 (2):\penalty0
  85--90, 2020.

\bibitem[Yankoski et~al.(2021)Yankoski, Scheirer, and
  Weninger]{yankoski2021meme}
Michael Yankoski, Walter Scheirer, and Tim Weninger.
\newblock Meme warfare: Ai countermeasures to disinformation should focus on
  popular, not perfect, fakes.
\newblock \emph{Bulletin of the Atomic Scientists}, 77\penalty0 (3):\penalty0
  119--123, 2021.

\end{thebibliography}

\end{document}